# Advanced Text Analytics - Graph Neural Network for Fake News Detection in Social Media

**Anantram Patel[a] and Vijay Kumar Sutrakar[b]**

[a]Scope College of Engineering, Bhopal, India, anantpatel.chh@gmail.com
[b]Aeronautical Development Establishment, Defence Research and Development Organisation, Bangalore, India, vks.ade@gov.in

**Abstract**

Traditional Graph Neural Network (GNN) approaches for fake news detection (FND) often depend on auxiliary, non-textual data such as user interaction histories or content dissemination patterns. However, these data sources are not always accessible, limiting the effectiveness and applicability of such methods. Additionally, existing models frequently struggle to capture the detailed and intricate relationships within textual information, reducing their overall accuracy. In order to address these challenges Advanced Text Analysis Graph Neural Network (ATA-GNN) is proposed in this paper. The proposed model is designed to operate solely on textual data. ATA-GNN employs innovative topic modelling (clustering) techniques to identify typical words for each topic, leveraging multiple clustering dimensions to achieve a comprehensive semantic understanding of the text. This multi-layered design enables the model to uncover intricate textual patterns while contextualizing them within a broader semantic framework, significantly enhancing its interpretative capabilities. Extensive evaluations on widely-used benchmark datasets demonstrate that ATA-GNN surpasses the performance of current GNN-based FND methods. These findings validate the potential of integrating advanced text clustering within GNN architectures to achieve more reliable and text-focused detection solutions.

*Keywords:* Deep Learning, Fake news detection (FND), Graph Neural Network (GNN), Topic Modelling, Word clustering (WC)

## 1. Introduction

Now a days, internet resources mainly, social media platforms (SMPs) have become a popular source for sharing and receiving information due to their accessibility and ease of use. It has also become one of the prone areas for spreading misinformation very rapidly. The vast number of users and the constant influx of new content make it nearly impossible to manually monitor and verify the accuracy of every message [1]. Misinformation poses serious societal risks, including increasing public anxiety, creating social divisions, and undermining democratic values [2-4]. This highlights the critical need for effective systems to detect and stop the mass spread of fake news on internet. GNN has revealed substantial promise in FND by integrating text information with structural relationships [5- 9]. Despite their success, most existing GNN-based approaches rely heavily on supplementary data, such as the dynamics of information propagation (e.g., retweets, endorsements, and comments) or connections within social networks, to construct graph representations. In scenarios where only textual data is available such as when a news article is first published these methods often fail to deliver effective results.

There are multiple advantages for relying solely on textual data for FND on varieties of SMPs. Text is the utmost prevalent and easily accessible form of information across platforms, enabling large-scale analysis without requiring complex or resource-intensive data collection methods. This approach also alleviates privacy concerns by avoiding the use of sensitive user information, such as personal network data or online interactions [10-11]. Moreover, analyzing text independently allows for the early identification of fake news, potentially preventing its engagement with users and the subsequent ripple effects in social networks. By addressing misinformation at its origin, this method can significantly limit its spread and societal impact. FND is often framed as a text classification problem, whereas GNNbased techniques have shown significant promise in the area of FND. Notable approaches, such as TextING [12], TensorGCN [13], HyperGAT [14], TextGCN [15] and TLGNN [16], apply GNNs to text classification tasks. These methods, however, typically use a single graph to represent the relationships between articles, which can limit the depth of information captured. Since textual correlations can manifest in multiple dimensions, relying on a single graph constrains the capability of a given model to fully capture complexity of these relationships. An approach is proposed in this paper by incorporating joint word and document clustering to build graphs that reflect the diverse semantic aspects of the text, enabling richer and more accurate representations.

A method based on GNNs, called Advance Text analysis Graph Neural Network (ATA-GNN) is proposed for more effectively leveraging textual information and capture sentimental knowledge for enhanced graph creation in FND. In this technique, several graphs have been created (that represent information from diverse viewpoints) by performing topic (cluster) modelling and selecting representative words from each topic. Utilizing different clustering configurations with varying cluster sizes further captures multi-dimensional text correlations. A rich, multi-faceted representation of article is getting generated by application of GNNs to these text-clustering graphs. In the present work, ATA-GNN is used against various benchmark datasets used by many researchers in the recent past. Results of ATA-GNN show consistently improved performance against other GNN-based methods, demonstrating the usefulness of topic modelling in improving FND.

## 2. Dataset

In the study, experiments are performed using datasets of Twitter15 [17], Twitter16 [17], and Pheme [18]. These datasets have a lot information, for example, source tweets (ST), userattributes (UA), retweet comments (RC), propagation structure (PS) of source tweet andretweet comments, and label for source tweet etc. In the present study, only ST and their labelare used for the FND. The proposed model is a binary classification system where tweets are categorized as either Fake or True. Any other labels present in the dataset are discarded. The model processes the STs through several pre-processing steps before they are used as input to the model. Words that are deemed irrelevant (stop words) (such as "the," "and", "is") areremoved. Non-alphanumeric symbols like punctuation are removed. Tweets with fewer than 3 words are removed due to their limited value for classification. All characters are converted to lowercase to standardize the text. Any URLs present in the tweets are discarded and in lastlemmatization (as text normalization technique) performed.

## 3. Previous work

The current models like GCAN [7], FANG [8], BiGCN [5], and UPFD [6], which generally leverage UA along with source tweets ST to augment the performance of classification model (refer Table 1 for further details). However, in the proposed approach, it intentionally omits UA from its analysis to reduce the complexity of data which leads to improved performance and lesser training time. This choice strengthens resilience in scenarios where user metadata like user name or origin of tweet is unavailable, limited, or privacy-restricted, enhancing its generalizability across diverse datasets and applications. Moreover, while the other available study in the current time [5, 9, 19] integrate RC to predict public sentiments and engagement patterns, the proposed model completely focuses on the originating ST. By concentrating on source tweets, model taps into the textual patterns and structural signals present in the initial post to assess news authenticity directly, sidestepping potential biases or noise from subsequent user comments. This streamlined approach not only reduces dependency on user-specific data but also allows ATA-GNN to perform effectively even in early detection scenarios, where only the initial content is available. Some of the other studies like FANG [8], where models' data is used as heterogeneous graph (HetG) to incorporate diverse node and edge types, TCGNN [1] adopts a homogeneous graph (HomG) structure. This approach prioritizes computational efficiency and simplicity by avoiding the complexity associated with managing multi-typed entities and relationships. While HetG can capture rich interactions, they often introduce significant computational overhead without guaranteeing performance gains. Furthermore, ATA-GNN sets itself apart from models like TextGCN [15] and HyperGAT [14], which are designed for text classification (TC), as well as NRGNN [20] and RSGNN [21], which focus on node classification (NC) (refer Table 1 for further details). Unlike these approaches, ATA-GNN is purpose-built for FND, leveraging a targeted feature extraction process to identify patterns and cues unique to misinformation. This tailored design enhances its precision and ensures its usefulness in finding fake news in diverse datasets.

**Table 1:** comparison with other state of art models

|  | GCAN [7] | NRGNN [20] | RSGNN [21] | FANG [8] | BIGCN [5] | GACL [19] | TCGNN [1] | ATA-GNN |
|---|---|---|---|---|---|---|---|---|
| Original Goal (OG) | FND | NC | NC | FND | FND | FND | FND | FND |
| Source Tweets (ST) | ✓ | ✓ | ✓ | ✓ | ✓ | ✓ | ✓ | ✓ |
| User | ✓ | ✗ | ✗ | ✓ | ✓ | ✗ | ✗ | ✗ |
| User Attribute | ✓ | ✗ | ✗ | ✓ | ✓ | ✗ | ✗ | ✗ |
| Retweet Comment (RC) | ✗ | ✗ | ✗ | ✗ | ✓ | ✓ | ✗ | ✗ |
| Propagation Structure (PS) | ✓ | ✗ | ✗ | ✓ | ✓ | ✓ | ✗ | ✗ |
| Graph Type (GT) | HomG | HomG | HomG | HetG | HomG | HomG | HomG | HomG |

## 4. Proposed model

The main component of the proposed model is use of topic modelling to get insight (feature vector) of the document in different topics(dimension). Topic modelling is a statistical method used to uncover the hidden thematic structure in a collection of documents. It automatically groupdocuments into topics (clusters) based on the co-occurrence patterns of words, without requiring predefined labels or supervision. Unlike hard clustering, where a data point belongs to a single cluster, topic modelling is analogous to fuzzy clustering (soft clustering). Each document can belong to multiple topics, with varying probabilities or weights indicating its association with each topic.

In the proposed work, Linear Dirichlet Allocation ($LDA$) which is generative probabilistic model is used for topic modelling. It models the documents as a mixture of latent topics. It alsoassumes that documents are blends of topics and topics are distributions of words. Individual document consists of a distribution of topics (e.g., Document A might be 70% Topic 1 and 30% Topic 2) and individual topic is consideredas a distribution of words (e.g., Topic 1 might heavily feature words like "machine," "learning," and "algorithm"). The words that belong to a topic or the probability of words belonging into a topic are calculated on the basis of Gibbs sampling.

Each topic (cluster) is represented by graph. Hence, the number of chosen topics in $LDA$ is the number of graphs $G = \{G^i\}_i^c$, $c$ is the number of topics which indicates the number of topic equal to number of graphs. Each document of the corpus belongs to the entire topic but with different probability. Each document represents vertices in the graph but with different feature vector representation in each topic graph so the number vertices in eachtopic graph are equal and edge connection in graph are different, indicating that their graph structure are not same. There is a set of labels for each document either fake (0) or true (1). The set of unique word in the corpus is called *corpus-dictionary* which is denoted by $W$.

## 5. Topic modelling and graph construction

The first step is topic modelling (soft clustering). It groups the most important words regarding each topic. In the next step, out of grouped word in each topic,most weighted words fromeach topic is selected and less important word in each topic is remove. In the third step, these selected words regarding each topic used as the dictionary of particular topic. Subsequently, the $TFIDF$ vectorization for feature calculation for each node of topic graph node is applied. The last step calculates the similarity between the node by using the $cosine-similarity$ for edge creation. In this way, the graph node and edges are created for each topic. Figure 1 shows the steps involved for the graph creation.

### 5.1 Topic modelling

$LDA$ used for topic modelling [22]. Each *document* is denoted by *d* and the collection of documents (which is also known as *corpus*) denoted by *D*, where $D = \{d_1, d_2, d_3, ..., d_n\}$. Before applying $LDA$, each document is represented as vector. The vectorization is done by $TFIDF$ as $V = TFIDF(D, W)$, where $V$ is the collection of vectors for each document in the corpus and dimension of $V$ is $R^{d_n \times w}$. Number of topics is the hyper-parameter in the $LDA$ [22], which will need to be adjusted for different variation. Subsequently,$LDA$ is appliedusing $T = LDA(V, c)$, where $T$ is collection of importance (weight) of each word (indictionary) belongs to different topic. Each word will belong to each topic with different importance (weightage). The dimension of the $T$ is $R^{c \times W}$.

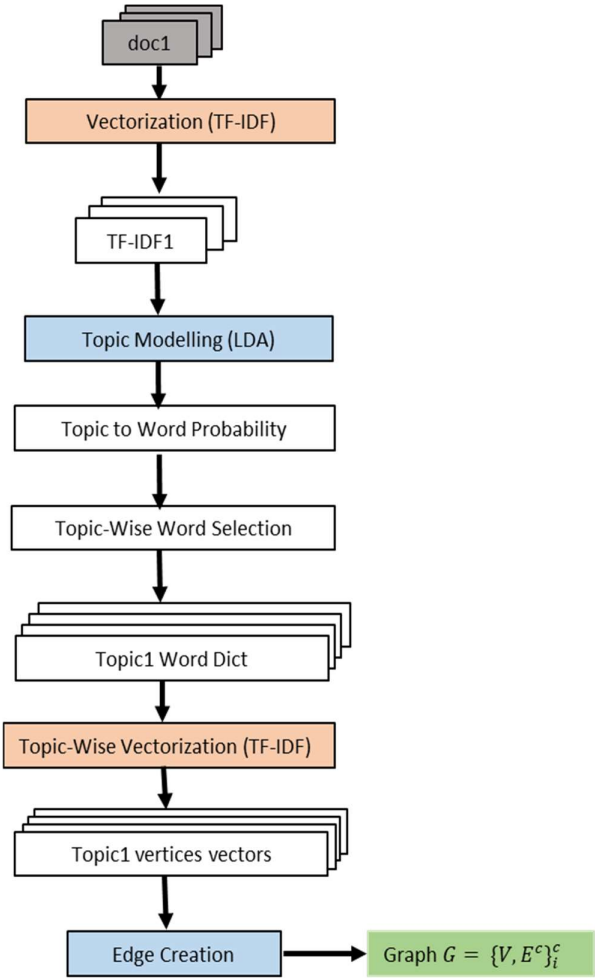

**Figure 1.** Steps involved in graph creation

## 5.2 Latent word selection in each topic

In the next step, the most representative words (which have the highest weight) in each topic is getting selected. Selected word in each topic will be treated as dictionary of that topic, which is called as *topic-dictionary and* denoted by $W_c$. The number of word selection out of $T$ in each topic is a hyper-parameter $r$ which varies between 0 and 1. When $r = 0$, means no word selected in the each topic and when $r = 1$ all the word selected in each topic when $0 < r < 1$, means number of words in each *topic-dictionary* will be more than zero and less than the number of word in $W$. Figure 2 shows top 10 selected word for one of the topic with their weightage(importance) in the topic. The $W_c = \{W_1, W_2, W_3, \ldots\ldots\ldots, W_n\}$, where $W_1$ is the *topic-dictionary* belongs to topic-1 and so on. Topic-wise word collection (*TWC*) is the algorithm which will select important word from topic-*dictionary* for each topic. The new topic-dictionary will be created for each topic which will have less word. The selection of most representative word will help in reducing the feature vector size because in *TFIDF* feature vector size is equal to length of dictionary.

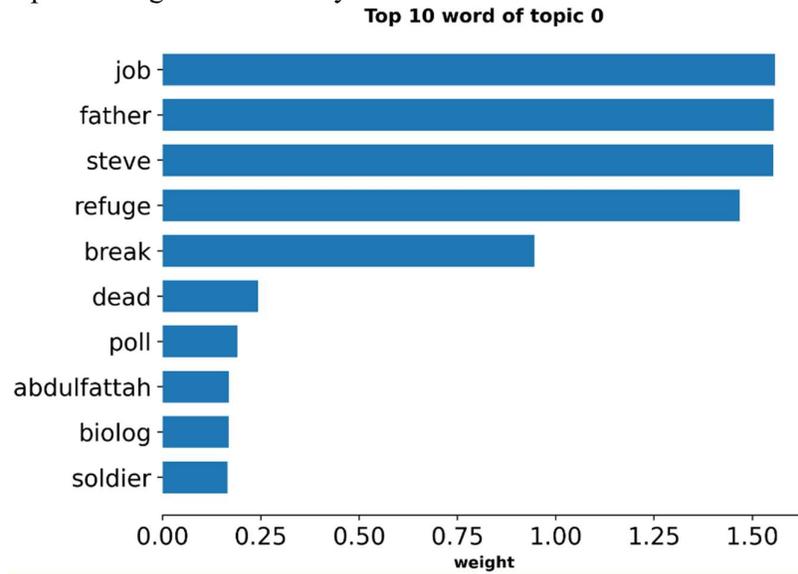

**Figure 2.** A typical case of top selected word with their weight

5.3 **Documents feature vector determination**

In the previous step, *TWC*, *topic-dictionary* for each topic is created. Each document belongs to a given topic and each topic has its own *topic-dictionary*. Each document is then transformed into the initial feature vector. Subsequently, one graph for each topic is created, called as *topic-graph*. In the graph, vertices represent the document. Since the number of documents in each topic is equal, the number of nodes in each *topic-graph* will also be equal. Again, *TFIDF* is used to create the initial feature vector of documents in the topic, $X_c = TFIDF(D, W_c)$, where $X_c$ is the feature-matrix of topic $c$ and $W_c$ is *topic-dictionary* belongs to topic $c$. In this way, each document becomes part of all topics; however, with different initial feature vector.

5.4 **Edge creation between documents**

In this step, the edges between the vertices (document) in each topic is created. For each topic c, edges are determined between vertices $v_c^i$ and $v_c^j$ by the help of cosine similarity. $v_c^i$ and $v_c^j$ are the initial feature vector of document i and j in the topic c. the cosine-similarity score will tell that one document in the topic how much similar to other documents in the same topic. cosine similarity is calculated on the basis of equation 5. Out of all edges of document, only *top-K* edges are selected based on the basis of similarity score.

6. **Model training**

In topic modelling, when each document is assigned to different topic with distinct initial feature vector, GNN gets different semantic and contextual meaning of the document. This is helping the model to analyze the document in the broader contextual aspect. The dataset is split into the ratio of 90:10. In the proposed model, there are 2 GCN layers(details are shown in Figure 3) with the internal dimensions of 64 and 32, respectively. ReLU function is used to capture the non-linearity of data in the hidden layer. After passing the GCN, there will be output embedding vectors to all documents with respect to each local graph. Subsequently, these embedding vectors are concatenate. After concatenating, there will be the final embedding $h_e$, where $h_e = h_{c2}^i \oplus h_{c2}^i \oplus h_{c4}^i \oplus h_{c4}^i \oplus h_{c4}^i \oplus h_{c4}^i \oplus h_{c4}^i$, and $\oplus$ is the concatenation operator. Let's assume, $p$ is the number of output features of a document that form GCN network and there is $q$ number of documents in the corpus. If the number of cluster is $(c_1, c_2, c_3, \ldots, c_z)$ then the dimension of final embedding vector $h_e$ of a document will be $h_e \in \mathbb{R}^{(c_1+c_2+c_3+\cdots+c_z)\circ p}$. The dimension of output final feature matrix of GCN network is $F_e \in \mathbb{R}^{q \times (c_1+c_2+c_3+\cdots+c_z)\circ p}$. The final embedding feature matrix is then feed to the fully connected dense layer using SoftMax activation function. Embedding feature matrix has two neurons (as the problem is binary classification). Cross-entropy is used as optimization function to be minimized during the training. Adam optimizer is used during the training to update the weight of the model [23].

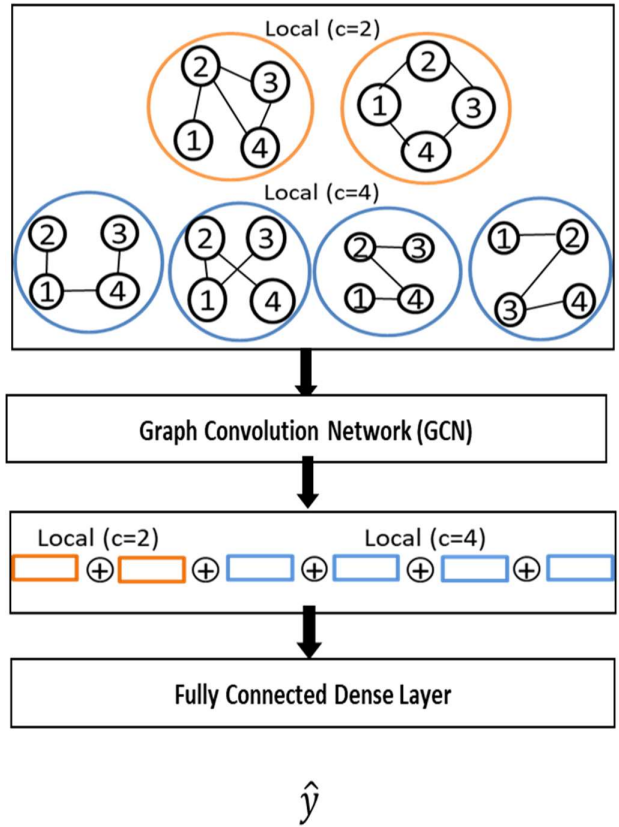

**Figure 3.** GNN flow

In the present work, fake news is labeled as 0 and true news is labeled as 1. If the model predicts the fake news as true news, then it will be more severe than model predict true new as fake new. It is due to the fact that fake news is very dangerous for any society. Hence, detection of fake news is more important than true news. In the study, precision is considered as an important metrics along with accuracy, recall and F1-score. The number of epochs used for the training are 300.

## 7. Results and discussions

Result of ATA-GNN model for the given datasets of Twitter-15 [17], Twitter-16 [17], and Pheme [18] is shown in Table – 2. Results of proposed model are also compared with the other state-of-the-art GNN based models for the FND in Table – 2. It can be seen that the proposed model outperformed in terms of accuracy, F1-score and Area Under Curve (AUC-score). The study is carried out for different cluster combinations of $H = \{\{8\}, \{8, 16\}, \{8, 16, 32\}\}$. The results of $H = \{8, 16, 32\}$ and *top-K* = 5 is shown in Table – 2. ATA-GNN shows accuracy of 90.20%, AUC of 96.27%, $F_1$-score of 90.23%, and Precision of 85.96% for Twitter-15 data [17]. ATA-GNN gives 6.74% higher accuracy, 7.37% higher AUC, and 0.5% higher $F_1$-score compared to state-of-the-art TCGNN model [1] for Twitter-15 data [17].

ATA-GNN shows accuracy of 92.73%, AUC of 97.60%, $F_1$-score of 93.19%, and Precision of 86.86% for Twitter-16 data [17]. ATA-GNN gives 7.70% higher accuracy, 8.11% higher AUC, and 3.06% higher $F_1$-score compared to state-of-the-art TCGNN model [1] for twitter16 dataset [17]. ATA-GNN shows accuracy of 91.54%, AUC of 93.61%, $F_1$-score of 92.68%, and Precision of 98.34% for Pheme data [18]. ATA-GNN gives % higher accuracy, % higher AUC, and % higher $F_1$-score compared to state-of-the-art TCGNN model [1] for Pheme data [18].

**Table 2:** ATA-GNN model prediction for Twitter15, Twitter16, and Pheme datasets

| | | GRU [24] | GraphSage [25] | Text-GCN [15] | RS-GNN [21] | NR-GNN [20] | TC-GNN [1] | ATA-GNN |
|---|---|---|---|---|---|---|---|---|
| Twitter15 | Accuracy | 0.7732 | 0.7537 | 0.7423 | 0.7477 | 0.7913 | 0.8450 | **0.9020** |
| | AUC | 0.7201 | 0.7837 | 0.4998 | 0.6702 | 0.0791 | 0.8966 | **0.9627** |
| | F1-Score | 0.8564 | 0.8532 | 0.5171 | 0.8554 | 0.8835 | 0.8977 | **0.9023** |
| | Precision | - | - | - | - | - | - | **0.8596** |
| Twitter16 | Accuracy | 0.7415 | 0.7537 | 0.7073 | 0.7707 | 0.7720 | 0.8503 | **0.9273** |
| | AUC | 0.6128 | 0.7837 | 0.5113 | 0.7295 | 0.0077 | 0.8949 | **0.9760** |
| | F1-Score | 0.8470 | 0.8532 | 0.5732 | 0.8664 | 0.8713 | 0.9013 | **0.9319** |
| | Precision | - | - | - | - | | | **0.8686** |
| Pheme | Accuracy | 0.8321 | 0.8338 | 0.6269 | 0.7906 | 0.8138 | 0.8672 | **0.9154** |
| | AUC | 0.8662 | 0.9070 | 0.5025 | 0.8660 | 0.7955 | 0.9440 | **0.9361** |
| | F1-Score | 0.7323 | 0.7515 | 0.6470 | 0.6426 | 0.6983 | 0.8188 | **0.9268** |
| | Precision | - | - | - | - | - | - | **0.9834** |

Next, the effect of *H*, for three different cases, $\{\{8\}, \{8, 16\}, \{8, 16, 32\}\}$, is shown. It can be clearly seen that as the number of cluster combination increased, there is an improvement in different comparison metrics like accuracy, F$_1$-Score, and AUC-score, as shown in Figure 4. The improvement in accuracy, F$_1$-Score, and AUC-score is observed in all the datasets, i.e. Twitter-15 [17], Twitter-16 [17], and Pheme [18], as shown in Figures 4(a), 4(b), and 4(c), respectively. The training time comparison is also shown in Figure 4(d). It is observed that training time grows almost exponentially (in general) as the number of cluster combination increased for a constant number of documents (sample) in the training data. Hence, judicious choice between the *H* and accuracy, F$_1$ score, and AUC score to be considered for optimal utilization of computational resources and optimal prediction of model performance.

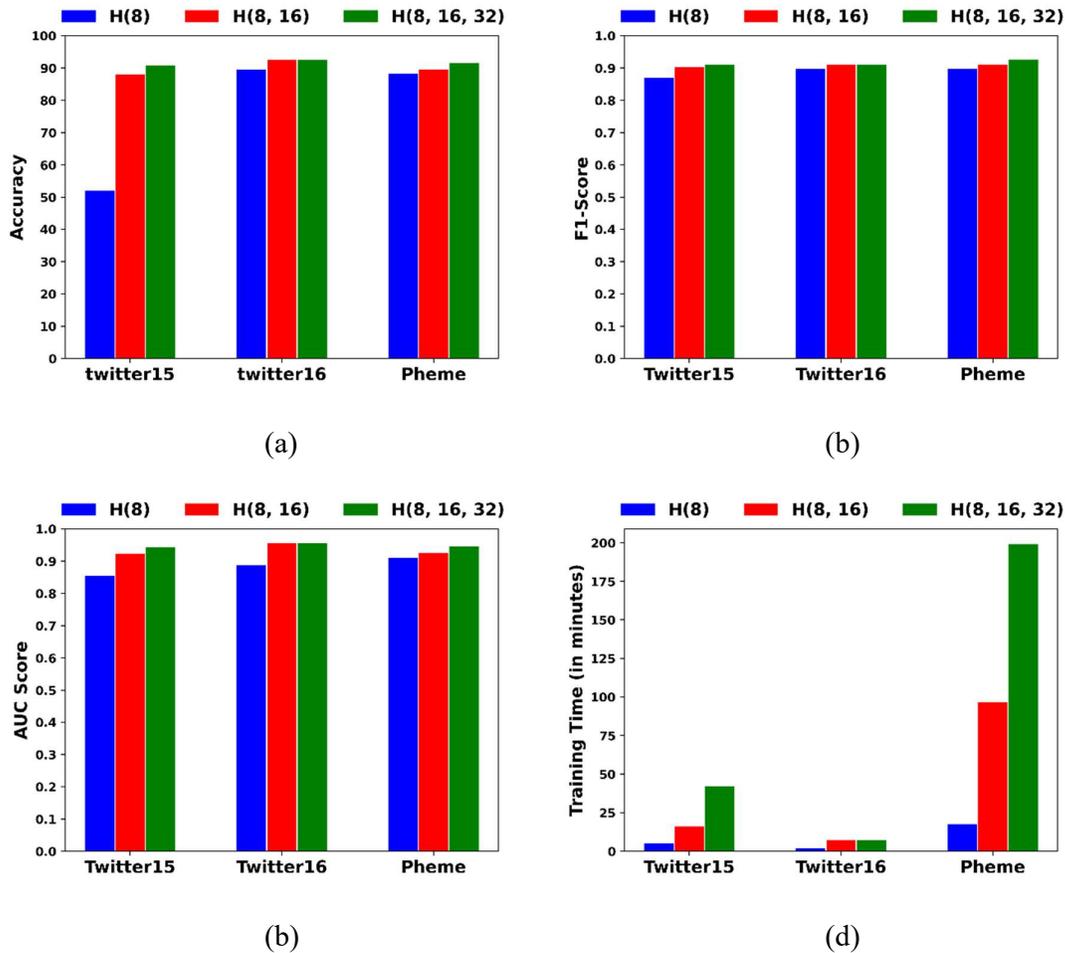

**Figure 4.** accuracy, F1-score, AUC-score and training time comparison

Lastly, the impact of *top-K* similarity is studied. It is one of the most important hyper-parameters for getting the information of the edge between the nodes. The only edge with highest similarity score is considered in the proposed model. It can be seen that initially the number of edges per node is increases and lead to improve F$_1$-score. However, too many edges per node lead to decrease in F$_1$-score, as shown in Figure 5 for Twitter-15 [17], Twitter-16 [17], and Pheme [18]. This is mainly due to the fact that as the number of edges increases, the neighbour node bringing the noise to the data rather than important information to the training data. Hence, it is also necessary to optimize the *top-K* value for different dataset.

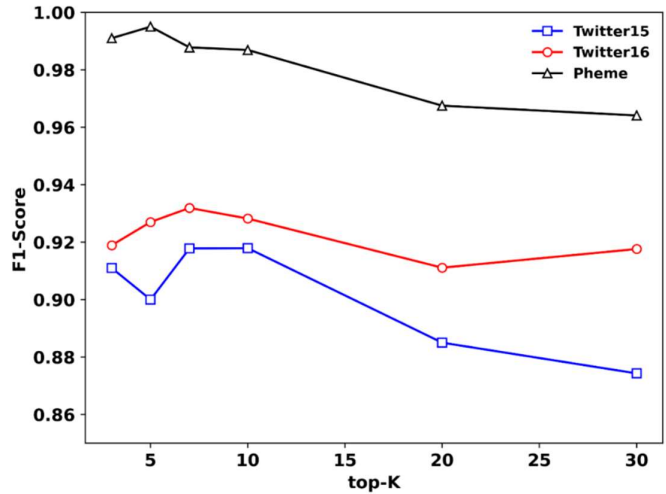

**Figure 5.** Variation of F1 Score with different *top-K*

## 8. Conclusions

In this work, an approach of Advance Text Analysis Graph Neural Network (ATA-GNN) is developed for fake news detection on the social media related graph-based data. This approach is distinct in term of graph construction from prior available study on the graph-based model for FND. Proposed model uses an efficient topic modelling-based rule that is based on soft clustering where each document can belong to all topics with distinct importance (weightage). In this setting, we outline the importance of topic modelling and creating multiple graphs for the corpus with different initial feature vector and distinct edge connection between the vertices of graph. The proposed study does not focus on the social media posts related information like location of user, name of user, and comments on the post and propagation structure of the post through social media; however, complete focus is given on the text data of a given post to create graph. ATA-GNN shows higher accuracy, higher AUC, higher $F_1$-score, and Precision compared to state-of-the art TCGNN model [1] for Twitter-15 [17] Twitter-15 [17], and Pheme [18] dataset. The word tokenization and vectorization technique can further be replaced with large language model (LLM) based tokenization and embedding technique in future, to get initial feature vector for the document. It may help to capture more contextual information about the text in the document and may further improve the efficiency of the model.

## References

1. Pei-Cheng Li, Cheng-Te Li. "Chapter 11 TCGNN: Text-Clustering Graph Neural Networks for Fake News Detection on Social Media", Springer Science and Business Meddia LLC, 2024
2. Bessi, A., et al.: Viral misinformation: the role of homophily and polarization. In: Proceedings of the 24th International Conference On World Wide Web, pp. 355–356 (2015)
3. Ribeiro, M.H., Calais, P.H., Almeida, V.A., Meira Jr, W. "EverythingI disagree with is #fakenews": Correlating political polarization and spread of misinformation. Data Science + Journalism (DS+J) Workshop @ KDD'17 (2017)
4. Vicario, M.D., Quattrociocchi, W., Scala, A., Zollo, F.: Polarization and fake news: early warning of potential misinformation targets. ACM Trans. Web (TWEB) 13(2), 1–22 (2019)

5. Bian, T., et al.: Rumour detection on social media with bi-directional graph convolutional networks. 34, 549–556 (2020)
6. Dou, Y., Shu, K., Xia, C., Yu, P.S., Sun, L.: User preference-aware fake news detection. In: Proceedings of the 44th International ACM SIGIR Conference on Research and Development in information Retrieval, pp. 2051–2055. SIGIR '21 (2021)
7. Lu, Y.J., Li, C.T.: GCAN: graph-aware co-attention networks for explainable fake news detection on social media. In: Proceeding of the 58th Annual Meeting of the association for Computational Linguistics, pp. 505–514 (2020).
8. Nguyen, V.H., Sugiyama, K., Nakov, P., Kan, M.Y.: Fang: leveraging social context for fake news detection using graph representation. In: Proceedings of the 29th ACM International Conference on Information and Knowledge Management, pp. 1165–1174 (2020)
9. Wei, L., Hu, D., Zhou, W., Yue, Z., Hu, S.: Towards propagation uncertainty: edge-enhanced Bayesian graph convolutional networks for rumor detection. In: Proceedings of the 59th Annual Meeting of the Association for Computational Linguistics, pp. 3845–3854
10. Kipf, T.N., Welling, M.: Semi-supervised classification with graph convolutional networks. In: International Conference on Learning Representations (ICLR) (2017)
11. Lian, Z., Zhang, C., Su, C., Dharejo, F.A., Almutiq, M., Memon, M.H.: Find: privacy-enhanced federated learning for intelligent fake news detection. IEEE Transactions on Computational Social Systems (2023)
12. Zhang, Y., Yu, X., Cui, Z., Wu, S., Wen, Z., Wang, L.: Every document owns its structure: Inductive text classification via graph neural networks. In: Proceedings of the 58th Annual Meeting of the Association for Computational Linguistics, pp. 334–339 (Jul 2020)
13. Liu, X., You, X., Zhang, X., Wu, J., Lv, P.: Tensor graph convolutional networks for text classification. In: Proceedings of the AAAI Conference on Artificial Intelligence, vol. 34, pp. 8409–8416 (2020)
14. Ding, K., Wang, J., Li, J., Li, D., Liu, H.: Be more with less: hypergraph attention networks for inductive text classification. In: Proceedings of the 2020 Conference on Empirical Methods in Natural Language Processing (EMNLP), pp. 4927–4936 (2020)
15. Yao, L., Mao, C., Luo, Y.: Graph convolutional networks for text classification. In: Proceedings of the AAAI Conference on Artificial Intelligence, pp. 7370–7377 (2019)
16. Huang, L., Ma, D., Li, S., Zhang, X., Wang, H.: Text level graph neural network for text classification. In: Proceedings of the 2019 Conference on Empirical Methods in Natural Language Processing (EMNLP), pp. 3444–3450 (2019)
17. Ma, J., Gao, W., Wong, K.F.: Rumour detection on Twitter with tree-structured recursive neural networks. In: Proceedings of the 56th Annual Meeting of the Association for Computational Linguistics, pp. 1980–1989 (2018)
18. Buntain, C., Golbeck, J.: Automatically identifying fake news in popular twitter threads. In: 2017 IEEE International Conference on Smart Cloud (smartCloud), pp. 208–215. IEEE (2017)
19. Sun, T., Qian, Z., Dong, S., Li, P., Zhu, Q.: Rumour detection on social media with graph adversarial contrastive learning. In: Proceedings of the ACM Web Conference 2022, pp. 2789–2797. WWW '22 (2022)
20. Dai, E., Aggarwal, C., Wang, S.: Nrgnn: learning a label noise resistant graph neural network on sparsely and noisily labeled graphs. In: Proceedings of the 27th ACM SIGKDD Conference on Knowledge Discovery and Data Mining, pp. 227–236 (2021)

21. Dai, E., Jin, W., Liu, H., Wang, S.: Towards robust graph neural networks for noisy graphs with sparse labels. In: Proceedings of the Fifteenth ACM International Conference on Web Search and Data Mining, pp. 181–191 (2022)
22. Blei, D.M., Ng, A.Y., Jordan, M.I.: Latent dirichlet allocation. J. Mach. Learn. Res. 3(Jan), 993–1022 (2003)
23. Kingma, D.P., Ba, J.: Adam: a method for stochastic optimization. In: Proceedings of International Conference for Learning Representations (ICLR) (2015)
24. Chung, J., Gulcehre, C., Cho, K., Bengio, Y.: Empirical evaluation of gated recurrent neural networks on sequence modeling. In: NIPS 2014 Deep Learning and Representation Learning Workshop (2014)
25. Hamilton, W., Ying, Z., Leskovec, J.: Inductive representation learning on large graphs. In: Advances in Neural Information Processing Systems 30 (2017)